\documentclass[runningheads]{llncs}
\usepackage{graphicx}
\usepackage{amsmath,amssymb} 
\usepackage{color}
\usepackage{floatrow}
\usepackage{xcolor,soul}
\usepackage{subcaption}
\usepackage[width=122mm,left=12mm,paperwidth=146mm,height=193mm,top=12mm,paperheight=217mm]{geometry}

\definecolor{note}{rgb}{0.7,0.1,0.5}

\begin{document}
\pagestyle{headings}
\mainmatter
\def\ECCV18SubNumber{1832}  

\title{Learning Kinematic Descriptions using \textbf{SPARE}: \newline \bf{S}imulated and \bf{P}hysical \bf{AR}ticulated \bf{E}xtendable dataset
	} 

\titlerunning{ }

\authorrunning{ }

\author{Abhishek Venkataraman$^1$, Brent Griffin$^{1,2}$, Jason J. Corso$^{1,2}$}
\institute{$^1$ Robotics Institute, University of Michigan, Ann Arbor\\
	  $^2$Electrical Engineering and Computer Science, University of Michigan, Ann Arbor\\
	  \{abhven, griffb, jjcorso\} @umich.edu}

\maketitle
\long\def\throwaway#1{}
\begin{abstract}
Next generation robots will need to understand intricate and articulated objects as they cooperate in human environments.  To do so, these robots will need to move beyond their current abilities---working with relatively simple objects in a task-indifferent manner---toward more sophisticated abilities that dynamically estimate the properties of complex, articulated objects.
    To that end, we make two compelling contributions toward general articulated (physical) object understanding in this paper.  First, we introduce a new dataset, \textbf{SPARE}: \textbf{S}imulated and \textbf{P}hysical \textbf{AR}ticulated \textbf{E}xtendable dataset.  SPARE is an extendable open-source dataset providing equivalent simulated and physical instances of articulated objects (kinematic chains), providing the greater research community with a training and evaluation tool for methods generating kinematic descriptions of articulated objects.  To the best of our knowledge, this is the first joint visual and physical (3D-printable) dataset for the Vision community.  Second, we present a deep neural network that can predit the number of links and the length of the links of an articulated object.  These new ideas outperform classical approaches to understanding kinematic chains, such tracking-based methods, which fail in the case of occlusion and do not leverage multiple views when available.


\throwaway{
{\color{note}From abstract, it should be clear that there are no deep learning methods that currently build DH tables or provide kinematic descriptions from observations. Establish that it is well known that deep learning methods require datasets, and in the absence of a dataset in this space, it is not surprising that we have yet to see DL solution in this problem space. We address this by introducing SPARE,

establish 1) learning-based methods for generating kinematic objects 2) other methods are not yet successful 3) DL could be successful, but there is no dataset 4) we introduce a dataset, extendedable so there is no limit to number of trained parameters and 5) show that it is viable for getting kinematic descriptions of objects
}}

\throwaway{
	{\color{note} (previous abstract)
		Low-cost learning is important given the complexity of our environments and the endless variation of objects people are expected to interact with. 
		In robotics, however, understanding new categories of objects typically requires the development of new processes of varying complexity, which is a significant barrier to general application robots.
		To help make general object understanding more practical, we introduce a new dataset, \textbf{SPARE}: \textbf{S}imulated and \textbf{P}hysical \textbf{AR}ticulated \textbf{E}xtendable dataset, and a new framework for learning the kinematic model of objects from RGBD video.
		By providing ground truth SPARE enables the training and evaluation of frameworks building kinematic models to match observed manipulation components and movement sequences, which is achieved by learning the number of links, number of DOF, relative link lengths, and relative link positions throughout the sequence.
		To demonstrate the efficacy of our approach, we teach our network to generate Denavit-Hartenberg parameters for simulated and physical SPARE objects.
	}
}
\end{abstract}

\section{Introduction}
%


With the success of robots in automation, their application is moving from controlled industrial environment to unstructured environments with unknown objects. 
In order to adapt to this environment change, computer vision must enable robots to observe and understand properties of new objects.
Humans perform well in unstructured environments, in part because they observe, learn, and understand objects through a more comprehensive set of properties. 
It is clear that working in such environments, requires more than just pixel-level semantic scene understanding. 
In this regard, Yuan et al. \cite{yuan2017connecting} encode physical properties of fabric by associating them with information from RGB, depth and high resolution touch sensors. 
Wu et al. \cite{WuYiLiJoFeTe15} predict mass and coefficient of friction of objects from videos using their physical scene-understanding model, Galileo.

\begin{figure}[t]
    {\caption{Human Support Robot(HSR) observing a physical SPARE dataset object. Because the HSR is mobile, it can collect multiple views of objects to learn their kinematic descriptions. HSR observes the object in the same pose over 5 views. }\label{fig:HSR}}
	{\includegraphics[width=7cm]{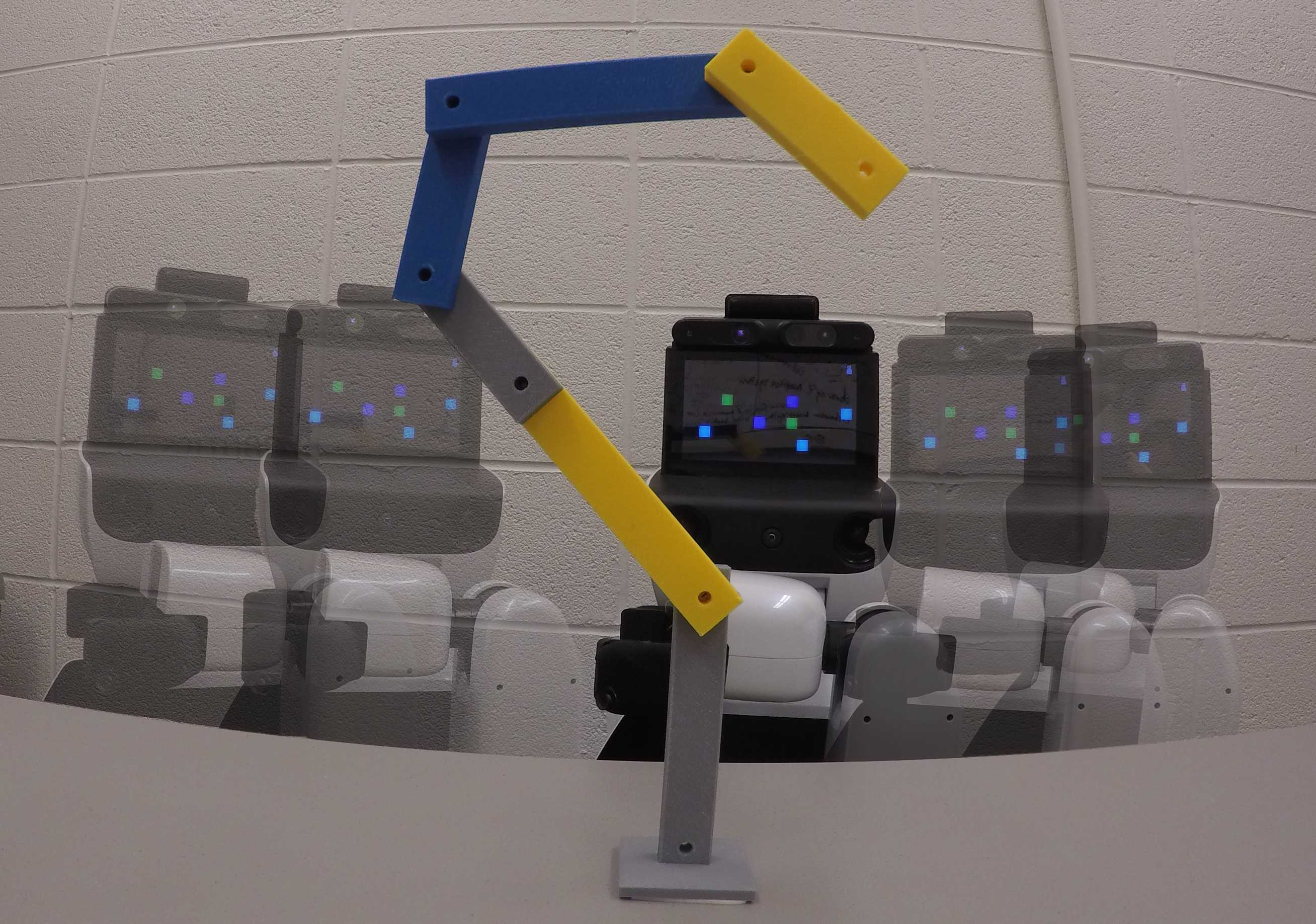}}
\end{figure}                                                                                     

In this work, we push further in this exciting direction to extend vision to include understanding of the kinematic properties of observed objects.
Whether innate or learned, it is clear that humans have a geometrical understanding of the many objects in our environment \cite{KeGuWiBOOK2001}.
We know that doors, for example, utilize a revolute joint hinged to a wall and, depending on the point of contact, require a variable amount of force to open---and \textit{we}, in a collective, human sense, know this even without any training or perhaps even if we do not have the appropriate words to express the geometry and functionality as we wrote in the antecedent.
Furthermore, not only are we good at manipulating known objects, we are also good at learning to understand new objects that we encounter~\cite{BeCOGN1995}. 
To work alongside humans in everyday, changing environments, robots must be able to do the same. 
More specifically, learning the kinematic descriptions of objects by observation is particularly useful for general manipulation in unstructured environments \cite{KaKaBaSt13}.



While there is relatively little work in generating kinematic descriptions for generalized articulated objects, a whole community has formed around estimating human pose in images and video \cite{YaPo08,PiAnGeSc13,YeYa14,VaDaMaJaYa14,ZhSuZhLiWeHuJe16}.
The cohesiveness of research in human pose estimation, in part, is due to the availability of datasets for this problem space \cite{JoEv10,SaTa13,IoPaOlSm14}.
However, \cite{JoEv10,SaTa13} have a limited number of training instances compared to the image- and video-based datasets typically required for training neural networks \cite{RuDeSuKrSaMaHuKaKhBeBeFe15,KaToShLe14,HeEsGhNi15}.
Deep learning-based human pose estimation such as \cite{ZhSuZhLiWeHuJe16}, use simulation-based dataset \cite{IoPaOlSm14}. 
Though human pose estimation requires encapsulation of articulation, understanding generic articulated object is a much harder problem space. First, in case of human pose estimation, the skeletal structure is available as prior. Second, the human pose estimation problem does not require exact angles of each joint.    


The more general methods for learning kinematic descriptions of articulated objects \cite{KaKaBaSt13,YaPo08,PiWaTe15,KuDhGaARXIV2016} have self-evaluated on closed datasets and focused on individual kinematic components rather than a complete description. While these tracking based methods have shown some success, their performance is severely affected when the links are under occlusion. 
With the recent success of deep neural network based approaches for image and video classification problems~\cite{KaToShLe14,RuDeSuKrSaMaHuKaKhBeBeFe15,Hi14} and robotics problems~\cite{valada2018deep,LeFiDaJMLR2016} there has been interest in extending these methods for other problems in robotics and vision. 
While, these methods have been quite successful, their performance is tied to the availability of large numbers of training examples. 
Annotating large datasets is generally a tedious task and often ends up being the bottle neck for learning larger networks.

In robotics, this requirement of large training examples makes learning-based methods difficult to apply on physical systems.
To overcome this limitation, recent work has focused on leveraging other tasks that have already been learned \cite{KrSu17} or transfer learning from simulation to real-world environments \cite{RuVeRoHePaHa16,ChShMoScBlToAb16,ToFoRaScZaAb17,MaBeHeScKrScTr17}.
For locating objects in complex scenes, \cite{ToFoRaScZaAb17} introduces variability during simulation-based training in an effort to make the real world appear as just another visual variation.
For robot manipulation tasks, \cite{ViPaSaPl17} trains networks in simulation to measure object distances from a wrist-mounted depth sensor on a UR5 robot, \cite{RuVeRoHePaHa16} grasps a block using progressive networks to transfer policies learned in simulation, and \cite{ChShMoScBlToAb16} learns to adapt the actions of a Fetch robot to compensate for friction and other physical discrepancies absent during simulation-based policy development.
In all cases, transfer-learning researchers are obligated to manually generate their simulation data and then attempt to replicate these training instances in a meaningful way for real-world implementation.

Cai et al \cite{Cai2017} have done a comprehensive study of current RGBD datasets that are acquired using Microsoft kinect or similar sensors. While these datasets are focused towards Human pose estimation, object detection/ recognition/ tracking or SLAM, none  of them have any annotated instances of general articulated objects. Thus there is a gap in this problem that can be addressed by introducing an annotated RGBD dataset for general articulated objects.



\begin{figure*}[t!]
	\centering 
	\includegraphics[width = 1\linewidth]{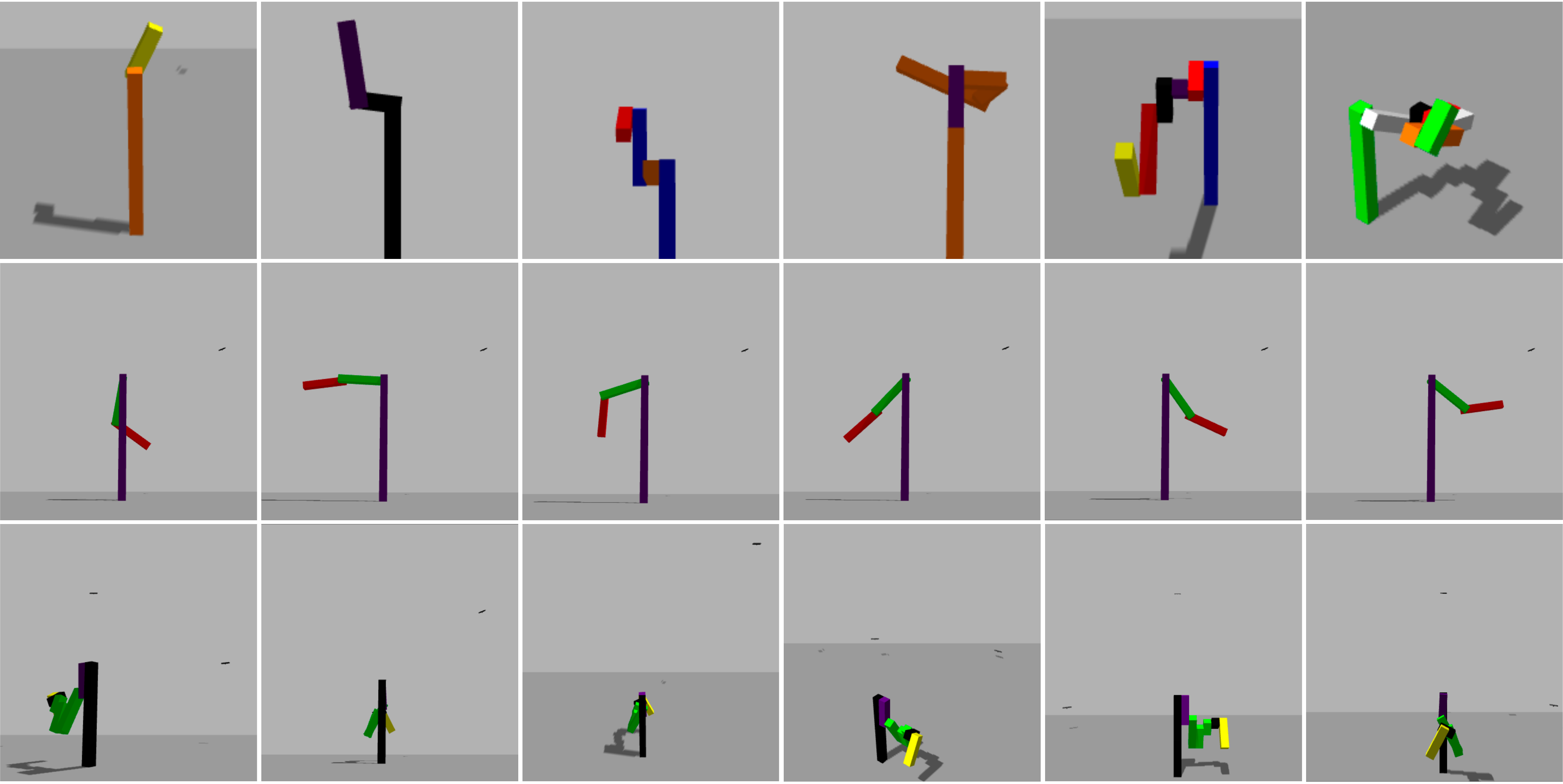}
	\caption{\label{fig:SPAREEx} Example SPARE images. The current SPARE implementation has kinematic chains consisting of one to six moving links (top row, ordered one to six), up to 100-frame temporal sequences (middle, spaced 20 frames apart), and up to eight simultaneous viewpoints (bottom, six views of a single time step). Corresponding depth images are not shown.}
\end{figure*}

%
In this paper, we introduce the SPARE: (\textbf{S}imulated and \textbf{P}hysical \textbf{AR}ticulated \textbf{E}xtendable) object dataset to provide a common training and evaluation tool for various methods generating kinematic descriptions of articulated objects (example object images are in Figure \ref{fig:SPAREEx}).
In addition, SPARE provides countless articulated objects in simulation with easily 3D-printable physical counterparts, which provides a direct link for transferring learning from simulation to real-world environments for observation- and manipulation-based tasks.
The simulation component provides RGB and depth image sequences, multiple views, and full 3D ground truth kinematic descriptions for dynamic articulated objects.
This variability enables the representation of many plausible robot-object interactions in simulation: a stationary robot watching moving objects, a mobile robot observing stationary objects (see Fig.~\ref{fig:HSR}), or even a team of robots making simultaneous observations from multiple view points.

\begin{figure}[h]
	\centering 
	\includegraphics[width = 0.6\linewidth]{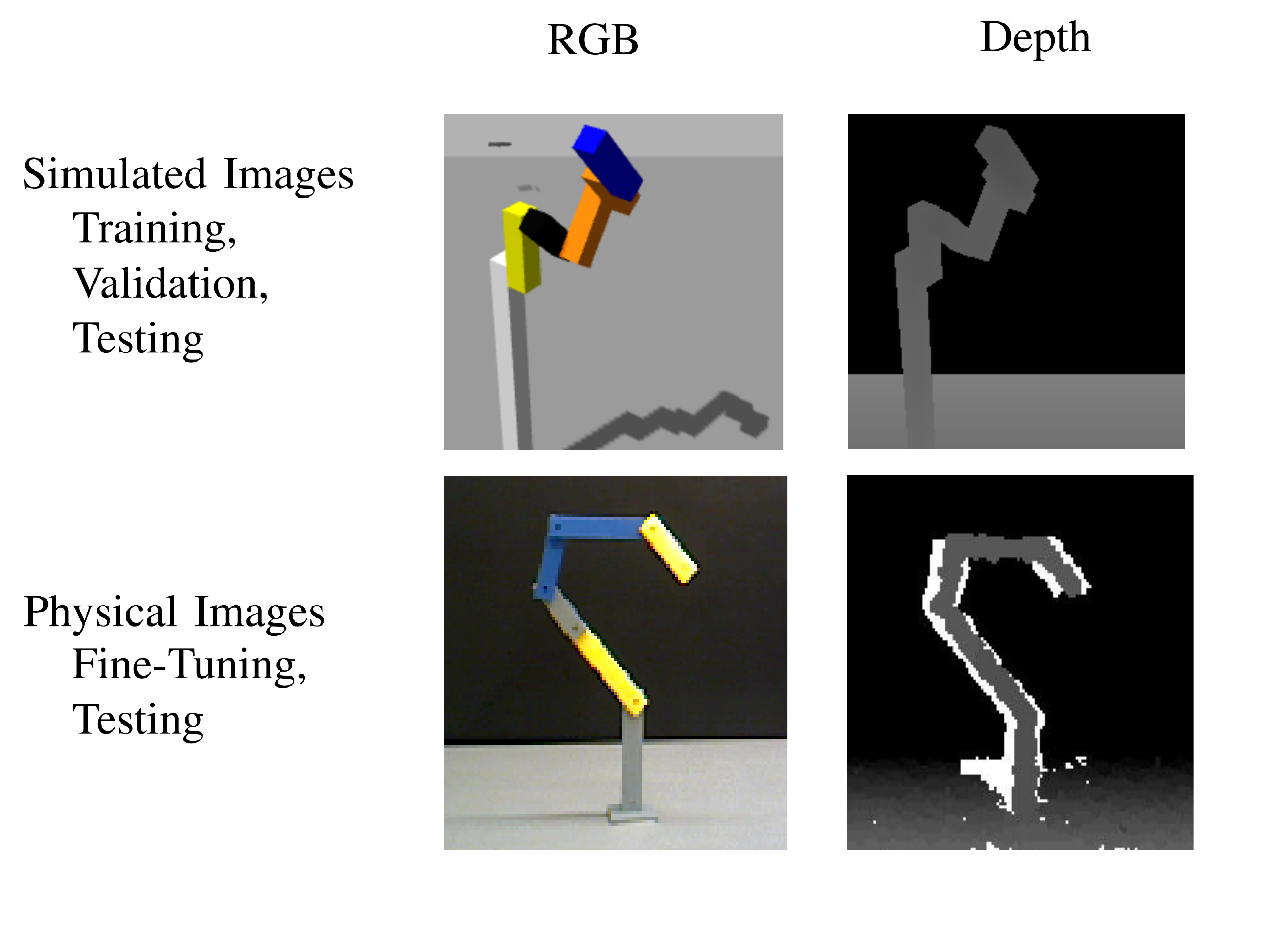}
    \caption{\label{fig:SPARE} The SPARE dataset contains kinematic chains that can be actualized in both simulated and physical environments. The open-source dataset is extendable in that users can generate additional randomized data as needed. }
\end{figure}

Furthermore, the dataset is open-source and extendable: if a user requires more data, SPARE can generate additional random instances of varying complexity to augment the original dataset (see Fig.~\ref{fig:SPARE}).
Given the introduction of a dataset with an extendable number of training instances, we train several deep neural network architectures to count the number of links in a kinematic chain and estimate the length of each link.
These architectures handle variation along critical axes: RGB versus depth, temporal versus single-frame, multi-view versus single-view.  Collectively, these architectures form a challenging, open-source baseline that, together with the SPARE dataset, lay the foundation to become the benchmark for future work in generating kinematic descriptions of articulated objects.
Our primary contributions are:
\begin{itemize}
	\item First, we introduce the SPARE object dataset, which, to our knowledge, is the first articulated object dataset to include physical and simulated components that can be extended as needed.  All code and (3D-printable) model files will be released with the paper.
	\item Second, we implement deep learning methods for identifying the number of links and regressing link lengths for static and dynamic objects in a scene, from either a single perspective or from multiple views. This constitutes the first link length estimation method for general articulated objects. 
	We envision it, along with the experimental setup, will become a benchmark for the community.
\end{itemize}

\section{\textbf{SPARE}: \textbf{S}imulated and \textbf{P}hysical \textbf{AR}ticulated \textbf{E}xtendable Dataset}\label{sec:SPARE}

%


In the space of articulated objects, there is no sufficiently large and diverse dataset available, to the best of our knowledge.  We hence propose the SPARE dataset.  SPARE not only provides annotated examples of articulated objects, but also allows the generation of addition examples. 
Thus, SPARE enables access to unlimited training examples, which makes it extremely relevant for approaches using deep neural networks.  Furthermore, these examples cross the simulation-physical boundary; for each articulated object defined in SPARE, simulation instances can be generated as needed \textbf{and} a physical 3D-printable object model specification is available.  We work with examples of both types in this paper demonstrating the power of crossing the simulation-physical boundary.

\begin{figure}[t]
	\centering 
	\includegraphics[width = 1\linewidth]{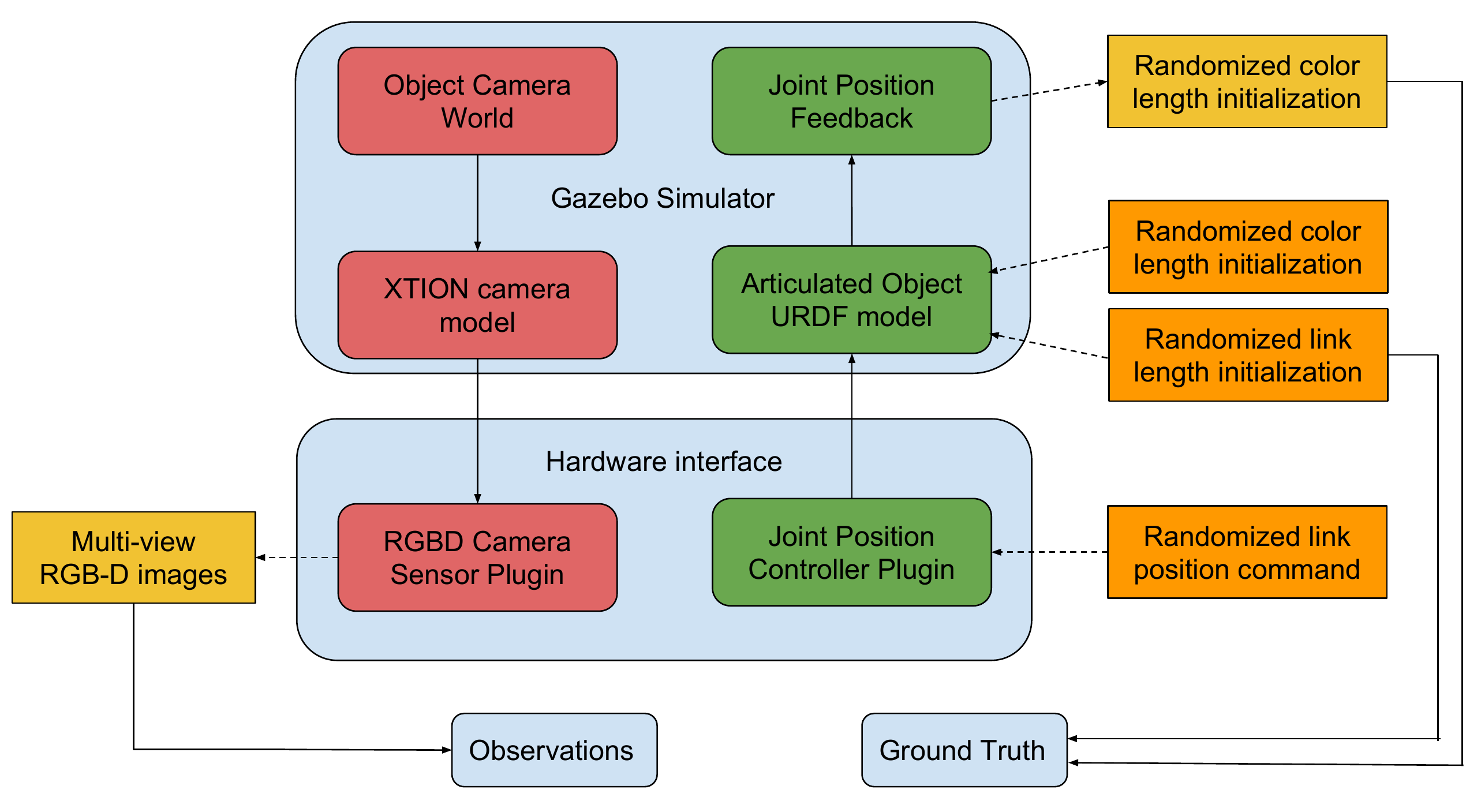}
	\caption{\label{fig:gazebo-methodology}Dataset generation flow-graph.  See text for details.}
\end{figure}

We gather the dataset by 
capturing RGBD images of various kinematic chains in the Gazebo physics simulation environment (Fig. \ref{fig:gazebo-methodology}; Sec. \ref{sec:framework}). 
Ground truth annotations for kinematic chains consist of the number of links, link lengths, and relative link positions throughout a dynamic sequence.
The dataset can be easily expanded by randomizing the parameters, such as number of links, color/lengths of links, etc. Additionally, the motion sequence of the links can be randomized, thus providing virtually infinite number of sequences with ground truth annotation. 
Through SPARE, the larger vision research community can train and evaluate different methodologies for building kinematic descriptions given observations of general articulated objects.

%

\subsection{SPARE Framework}
\label{sec:framework}

We generate simulated instances in SPARE using the Gazebo \cite{gazebo} simulation environment.
ROS provides the communication interface which allows sending joint control messages and receiving sensor data(images from RGBD cameras and joint angles).
The articulated object world consists of one multi-link object observed by 8 RGBD cameras from various viewpoints. 
 Fig.~\ref{fig:gazebo-methodology} describes the data flow for the creation of the dataset.  
Joint angle of each link in the object is controlled through the controller plugin. 
We achieve the object motion by updating the joint angles positions and then observing the object from 8 RGBD cameras.
In the current framework, we capture images at 10fps.


For each simulation instance, given the number of moving links, $n$, we create a configuration with random link lengths $\mathbf{L^{\prime}}$ and color set $\mathbf{C}$:
\begin{align}
L^{\prime}_i \sim &  \begin{cases}
U [1.3,2.0], ~\text{if}~ i==0 \\
U [0.3,1.0], ~\text{otherwise}
\end{cases} , ~i \in \{0,1,..,n\} 
\\
C_i \sim & U[\texttt{\scriptsize{black,white,red,orange,blue,green,yellow,indigo}}].
\nonumber
\end{align}

In this paper, we consider articulated objects that are structured as chains with links numbering zero through six. The zeroth link is the root link and is stationary. Each of the other links are connected to their parent link with a revolute joint.
Examples demonstrating the variation in the dataset are provided in Fig. \ref{fig:SPAREEx}.
In the case of longer chains, it is possible that sum of the lengths of the links is large causing the object to be outside the camera's viewing frustum. We address this problem with normalizing configurations with lengths greater than 3, such that total length of the link is equal to 3 as follows:
\begin{align}
\mathbf{L} = &  \begin{cases}
\mathbf{L}^{\prime}, ~\text{if}~ \Sigma L^{\prime}_i < 3  \\
\mathbf{L}^{\prime}*\dfrac{3}{\Sigma L^{\prime}_i}, ~\text{otherwise}
\end{cases} , ~i \in \{0,1,..,n\} .
\end{align}  
Throughout the SPARE dataset, we use the term \textit{number of moving links}, $n$, to refer to the actuated links; it does not include the base link. The term \textbf{number of links},  $n+1$, ; however,  includes the base link ($i=0$).


\subsection{Data Generation}
\label{sec:label_gen}

SPARE  can generate annotated instances when the number of moving links {($n$)} of object is specified. An object model is created with a random configuration of $\{ \mathbf{L,C} \}$ with {$n+1$} links. At the same time, the frames are annotated with ground truth information of the object properties, number of links and link lengths. A sequence of random motion is then generated and the observations from the 8 RGBD cameras is recorded. Figure \ref{fig:SPAREEx} shows the examples of various configuration and views of articulated objects in the dataset. The randomization of joint angles allows the links to transition between visible and occluded states.   

The generated image frames can be stacked according to the modality required. 
For applications which require temporal images, such as a stationary camera observing moving a object, sequential images from a single camera can be stacked together. 
For applications which require multiple views of a single object, such as a robot moving around a stationary object, images from multiple cameras at a single time step are stacked together. 
Multi-view is quite common in active perception; Eidenberger et al. \cite{EiScIROS10} perform planning for scene modeling in cluttered environment from 8 different view points.




\throwaway{
	\begin{figure}
		\begin{subfigure}[ht]{0.24\linewidth}
			\centering\includegraphics[width=1.0\linewidth]{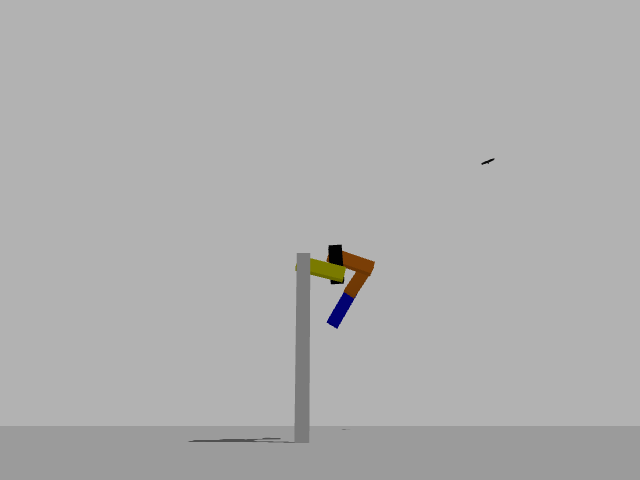}
		\end{subfigure}
		\begin{subfigure}[ht]{0.24\linewidth}
			\centering\includegraphics[width=1.0\linewidth]{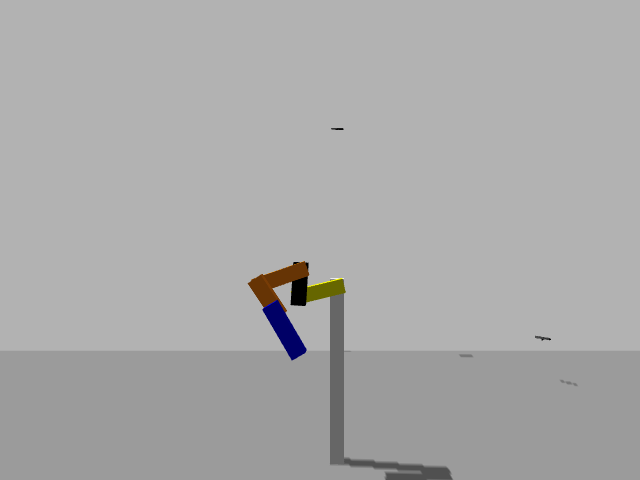}
		\end{subfigure}
		\begin{subfigure}[ht]{0.24\linewidth}
			\centering\includegraphics[width=1.0\linewidth]{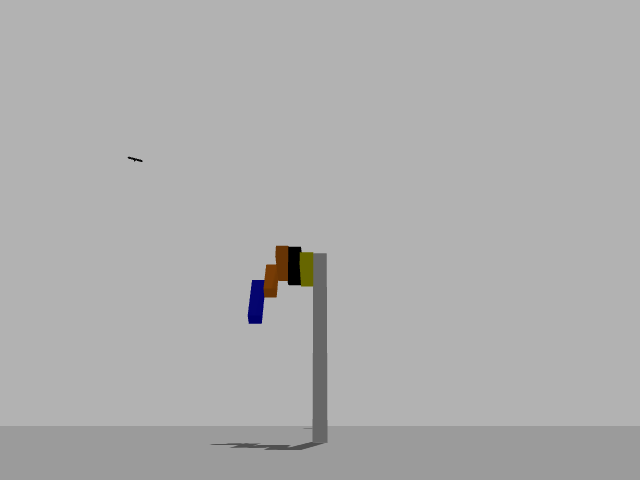}
		\end{subfigure}
		\begin{subfigure}[ht]{0.24\linewidth}
			\centering\includegraphics[width=1.0\linewidth]{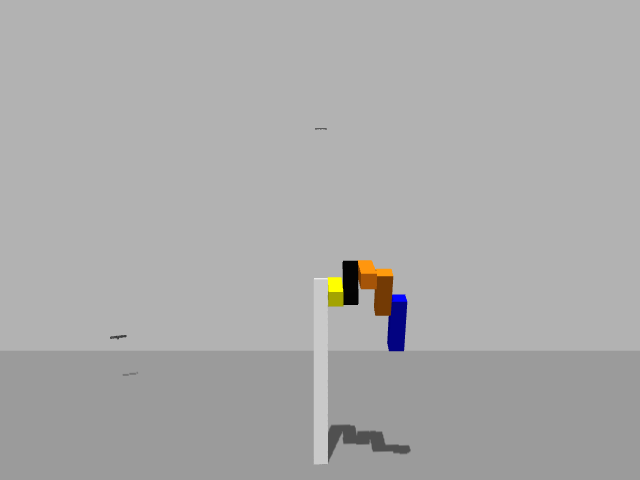}
		\end{subfigure}
		
		\vspace{1mm}
		
		\begin{subfigure}[ht]{0.24\linewidth}
			\centering\includegraphics[width=1.0\linewidth]{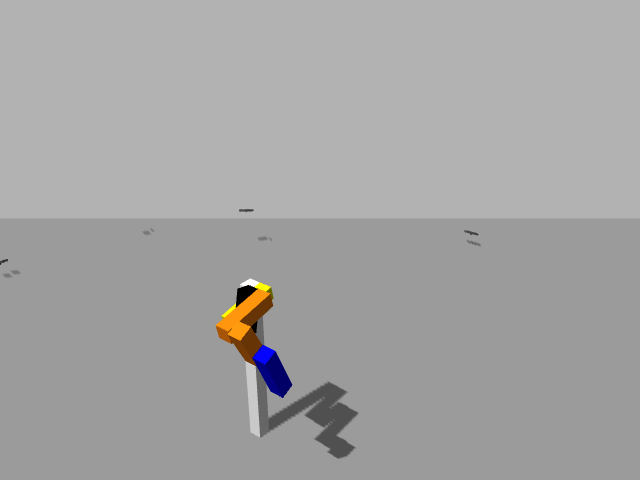}
		\end{subfigure}
		\begin{subfigure}[ht]{0.24\linewidth}
			\centering\includegraphics[width=1.0\linewidth]{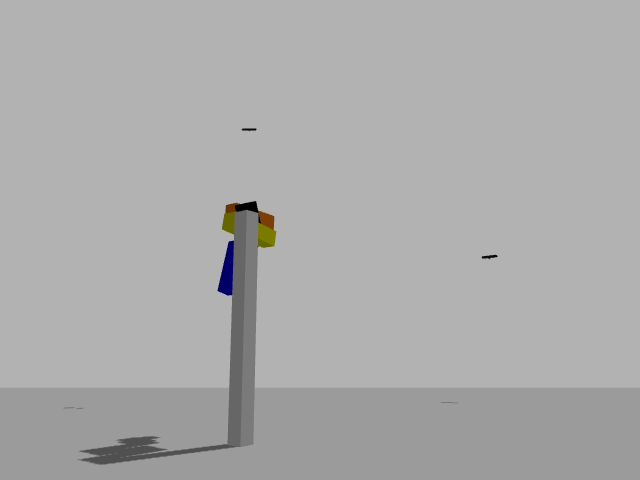}
		\end{subfigure}
		\begin{subfigure}[ht]{0.24\linewidth}
			\centering\includegraphics[width=1.0\linewidth]{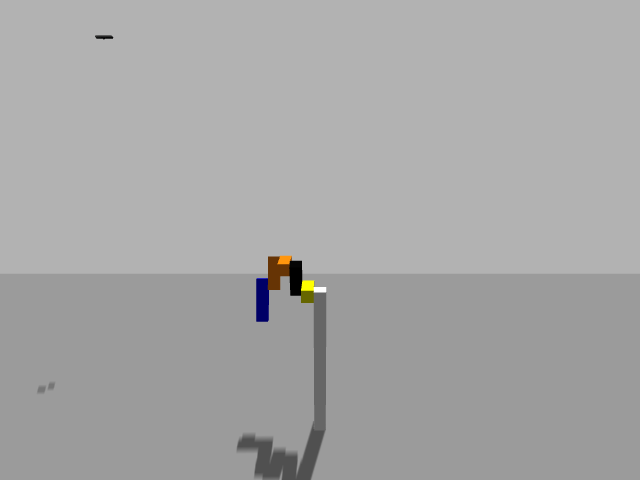}
		\end{subfigure}
		\begin{subfigure}[ht]{0.24\linewidth}
			\centering\includegraphics[width=1.0\linewidth]{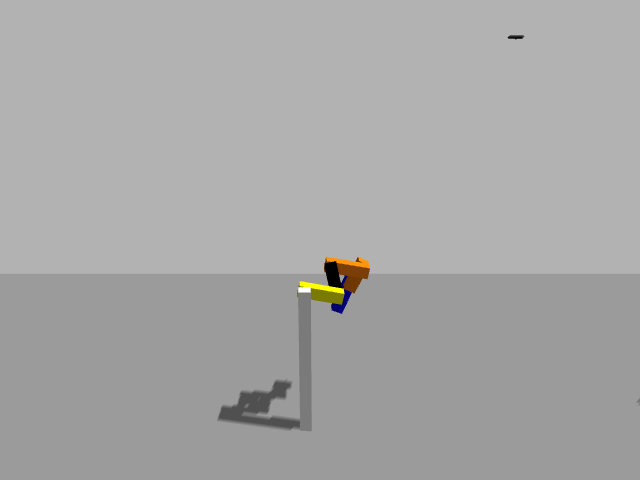}
		\end{subfigure}
		\caption{\label{fig:images-multiview}Multi-view RGB images from 8 RGBD (Xtion) cameras at a given instance \hl{I'd like more than one of these.  I'd like these to be zoomed in a bit more so that I can see what is happening.}}
	\end{figure}
	
	\begin{figure}
		\begin{subfigure}[ht]{0.24\linewidth}
			\centering\includegraphics[width=1.0\linewidth]{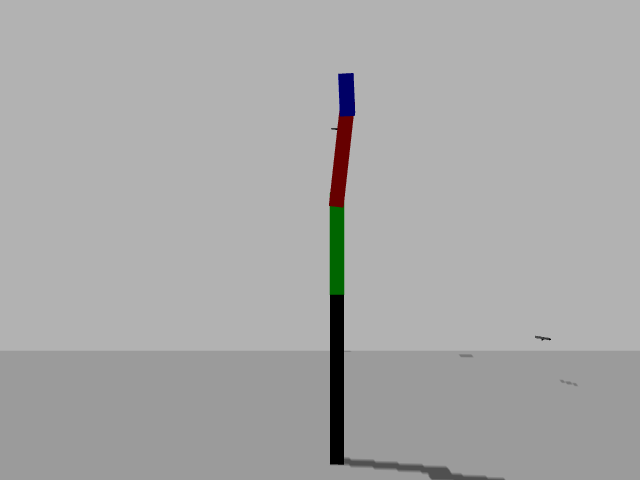}
		\end{subfigure}
		\begin{subfigure}[ht]{0.24\linewidth}
			\centering\includegraphics[width=1.0\linewidth]{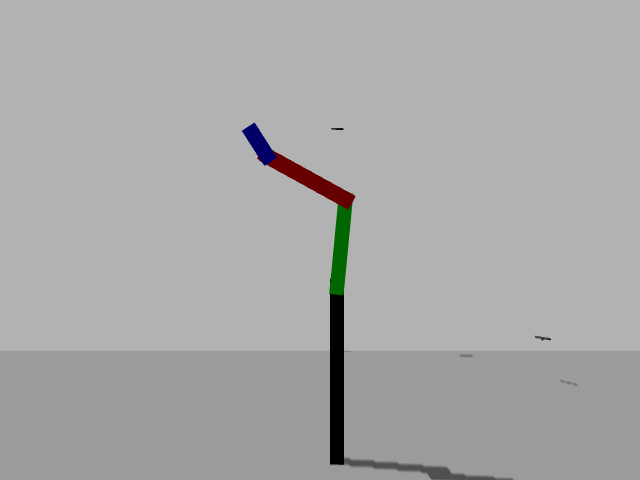}
		\end{subfigure}
		\begin{subfigure}[ht]{0.24\linewidth}
			\centering\includegraphics[width=1.0\linewidth]{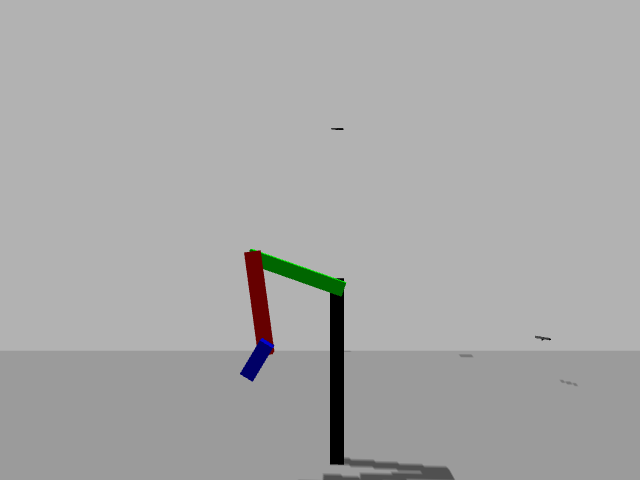}
		\end{subfigure}
		\begin{subfigure}[ht]{0.24\linewidth}
			\centering\includegraphics[width=1.0\linewidth]{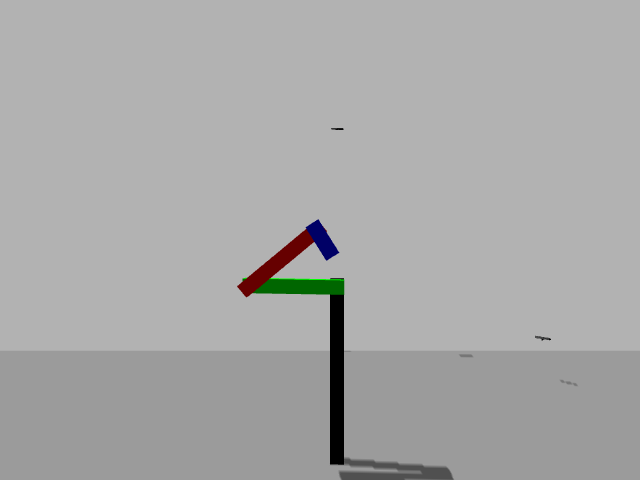}
		\end{subfigure}
		\caption{\label{fig:images-temporal}Temporal RGB images from a single XTION camera}
	\end{figure}
}

\section{Kinematics Learning Networks}  
\label{sec:ARCH}


In this section, we implement deep neural network architectures trained on SPARE to predict kinematic descriptions of objects. 
All the architectures we implement use raw rgb and depth image as input.
Given the scope of kinematic characteristics we are gathering from the dataset, we use multiple architectures, each designed for specific tasks. 
Namely, architectures counting the number of links in a scene are described in Section~\ref{sec:linkCount}, architectures estimating link lengths are described in Section~\ref{sec:linkLength}, and architectures simultaneously counting links and estimating link length are described in Section~\ref{sec:end_to_end}.


\subsection{Link Counter}
\label{sec:linkCount}

While we encounter many distinct articulated objects in the real world, most of them are restricted to a very small number of links. 
To put this in perspective, all of the following are two links or less: doors, faucets, scissors, pens, etc.
To account for these objects and those of increasing levels of complexity, the current SPARE implementation generates kinematic chains ranging from one to seven links.
This also allows us to estimate the number of links as a classification problem rather than a regression one.  

As discussed in Section~\ref{sec:SPARE}, depending on the modality of the robot's observation, the input stream can consist of either multiple views or temporal sequences.
Fig.~\ref{fig:counter_architectures} shows two architectures that perform well in predicting the number of links given multiple depth or gray-scale images as input.
The first architecture, 3D Convolution, allows stacking of multiple images.
Intuitively, this method spreads information spatially within a frame and across many frames.
Fig.~\ref{fig:counter_Con3D} depicts the architecture using 3D convolution.

The downside of 3D convolution is scalability; if the number of temporal frames or views increases, the number of trainable parameters in the network grows exponentially.
The second architecture avoids intractable parameters by 1) sharing convolution weights across multiple frames and 2) reducing images to a much smaller feature vector.
This smaller vector representation is then combined using LSTM layer \cite{HoSc97} with 64 hidden layers, as depicted in Fig.~\ref{fig:counter_LSTM}.

When using temporal images, both networks input 100 sequential frames of size 96$\times$128$\times$1 (gray-scale or depth images), and, when using mutli-view images, both networks input eight simultaneous views each of size 96$\times$128$\times$1 (gray-scale or depth image). 
Both output a six-dimensional vector in the final layer with softmax activation function.
The networks are trained end-to-end with a one-hot labels vector $v \in \mathbb{R}^6$ with a categorical cross entropy loss function.

\subsection{Link Length}
\label{sec:linkLength}
\begin{figure}
	\centering
	\begin{subfigure}[b]{0.35\linewidth}
		\includegraphics[width = 1\textwidth]{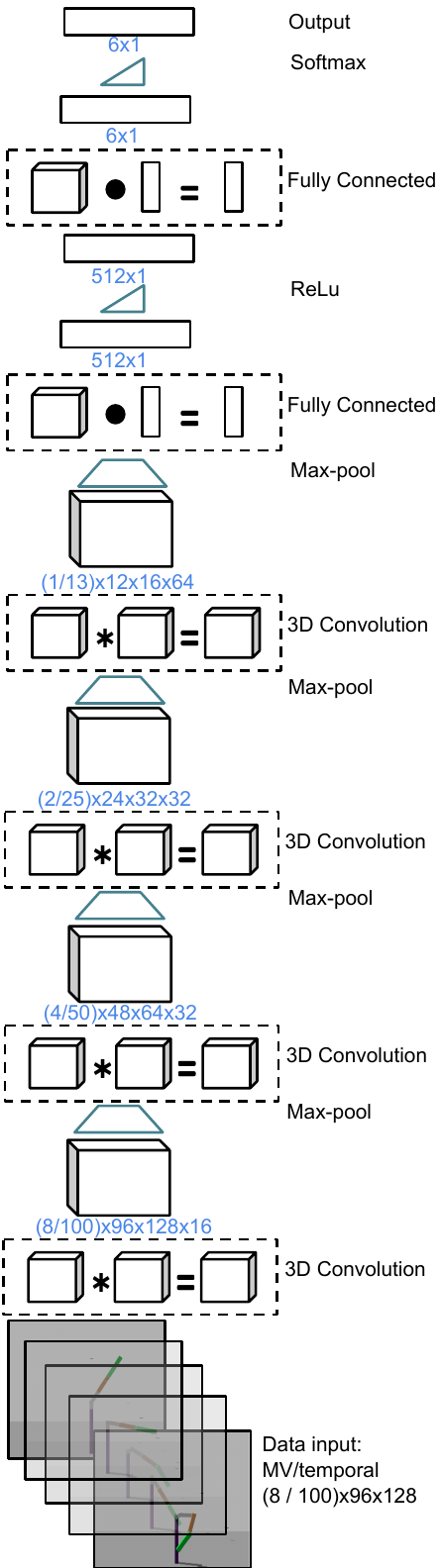}
		\caption{\label{fig:counter_Con3D}3D Convolution Network}
	\end{subfigure} 
	\hspace{8mm}
	\begin{subfigure}[b]{0.35\linewidth}
		\includegraphics[width = 1\textwidth]{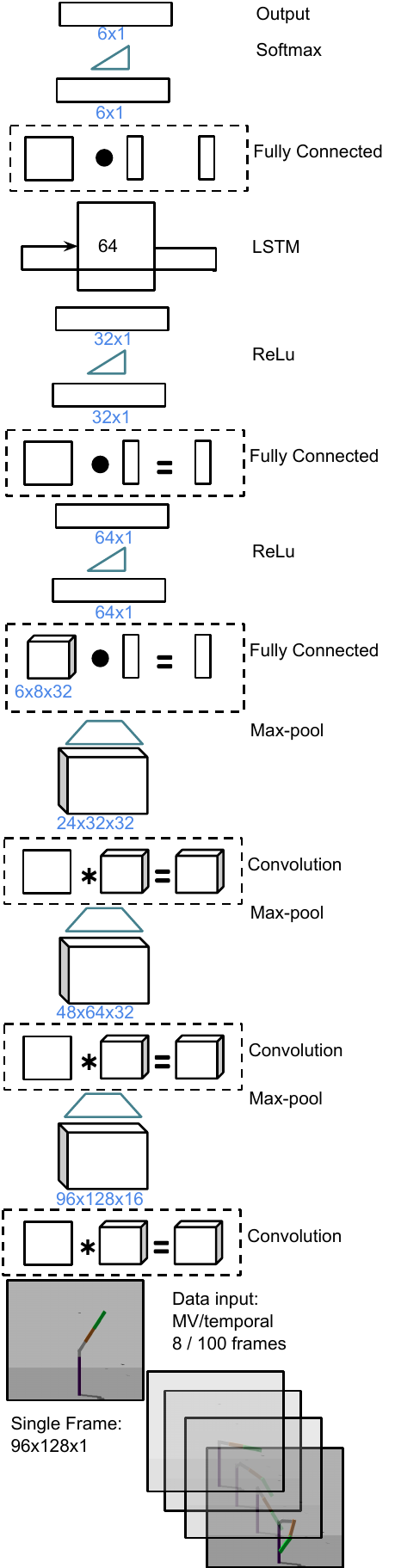}
		\caption{\label{fig:counter_LSTM}CNN-LSTM Network}
	\end{subfigure}
	\caption{\label{fig:counter_architectures}Schematic for 3D Convolution and CNN-LSTM link counting networks.}
\end{figure}

Given the number of links for an articulated object, the next step is to estimate the length of each link.
We use the 3D Convolution architecture again; however, unlike counting, estimation of link lengths is a regression problem.
To achieve this, we retain all four 3D Convolutional layers (see Fig.~\ref{fig:counter_Con3D}) and replace the two fully connected layers with the following three fully connected layers: 
\begin{itemize}
	\item FC1: 512 nodes ReLu activation
	\item FC2: 512 nodes, Linear activation
	\item FC3: \{{\# of moving links (n)}\} nodes, Linear activation
\end{itemize}
Note that the number of nodes in the last fully connected layer is variable and depends on the number of links estimated by the counting network.
Hence, we have six different networks (one for each link quantity), each training with link quantity-specific examples.

\setlength{\tabcolsep}{4pt}
\begin{table}[t]
\begin{center} 
\caption{SPARE evaluation and benchmark results.} 	
\label{tab:SPARERes}
\resizebox{\columnwidth}{!}{%
	\begin{tabular}{ l | l | c  c  c  c | c | c | c | c  c}
		\hline
		\multicolumn{1}{c|}{} & \multicolumn{1}{c|}{} & \multicolumn{5}{c}{} & \multicolumn{1}{|c}{} & \multicolumn{1}{|c}{} & \multicolumn{1}{|c}{} & \multicolumn{1}{c}{} \\
		
		\multicolumn{1}{c|}{} & \multicolumn{1}{c|}{Network} & \multicolumn{5}{c|}{Network Input Images}  &
		\multicolumn{1}{|c}{Training} & \multicolumn{1}{|c}{Test} & \multicolumn{2}{|c}{Evaluation} \\
		
		\cline{3-7} \cline{10-11}
		
		\multicolumn{1}{c|}{Architecture} & \multicolumn{1}{c|}{Evaluation}  & \multicolumn{1}{c}{Greyscale} & \multicolumn{1}{c}{Depth} &
		\multicolumn{1}{c}{Temporal} & \multicolumn{1}{c|}{Views} & \multicolumn{1}{c|}{Total} & \multicolumn{1}{c}{Instances} &
		\multicolumn{1}{|c}{Instances} & \multicolumn{1}{|c}{Accuracy} & \multicolumn{1}{c}{Error} \\
		\hline
		
		CONV3D-Depth-TMP & \# moving links & 0 & 1 & 100 & 0 & 100 & 1536 & 768 & 0.682 & \\
		LSTM-Depth-TMP      & \# moving links & 0 & 1 & 100 & 0 & 100     & 1536 & 768 & 0.638 & \\
		CONV3D-Depth-MV & \# moving links & 0 & 1 & 0    & 8 & 8     & 18000 & 9600 & \textbf{0.949} & \\
		LSTM-Depth-MV & \# moving links & 0 & 1 & 0    & 8 & 8     & 18000 & 9600 & \textbf{0.956} & \\
		CONV3D-Grey-TMP & \# moving links & 1 & 0 & 100 & 0 & 100     & 1536 & 768 & 0.559 & \\
		LSTM-Grey-MV & \# moving links & 1 & 0 & 0     & 8 & 8     & 18000 & 9600 & 0.891 & \\
		\hline
		
		CONV3D-Depth-TMP &  link lengths & 0 & 1 & 100    & 0 & 100 & 1496 & 240 &  & 6.64 \\
		CONV3D-Depth-MV &  link lengths & 0 & 1 & 0    & 8 & 8 & 17600 & 3000 &  & \textbf{0.543} \\
		\hline
		
		CONV3D-Depth-TMP & end-to-end & 0 & 1 & 100    & 0 & 100     & 1632 & 288 & & 14.8  \\
		CONV3D-Depth-MV & end-to-end & 0 & 1 & 0    & 8 & 8     & 25500 & 4500 &  &  \textbf{0.415}\\
		\hline
		
		
	\end{tabular}
}
\end{center}	
\end{table}
\setlength{\tabcolsep}{1.4pt}

\subsection{End-to-End Networks}
\label{sec:end_to_end}

Our end goal in the prediction process is to estimate the lengths of each link in the object. We achieve this end-to-end prediction using two architectures:
\begin{itemize}
\item[1)] \textit{Naive Combination:} \\
In this architecture, we combine the Link Length network with the Link Counter network to form a 2 stage end-to-end system.
We first train the 7 networks (1 Link Counter and 6 Link Length) individually. At prediction time, we first pass the test instance through the Link counter network and predict the number of moving links. We next choose the Link Length network based on this predicted output and use this to estimate the link lengths for the given test instance. \\

\item[2)] \textit{End-to-End trainable network:}\\ 
While the previous method is easy to implement, the combination network is not trained end-to-end.
Additionally, there is no knowledge or weights sharing between the six networks that estimating link-lengths.
Thus, to improve the performance, we implemented a single network that can be trained end-to-end and can work with any number of links.\\
To handle link-quantity variation, this end-to-end trainable network uses a modified link length label, $\mathbf{L} \in \mathbb{R}^7$, which is a seven-dimensional vector. 
Each component of the link length ground truth label, $L_i \in \mathbf{L}$, is calculated as:
\begin{align}
L_i =  \begin{cases}
l_i, ~\text{if}~ i<=n \\
0, ~\text{otherwise}
\end{cases}, ~i \in \{0,1,..,n\},
\label{eq:end-end}
\end{align}  
where $l_i$ is the ground truth length of link $i$, and {$n$} is the ground truth number of moving links in the instance object.
\end{itemize}

\subsection{Implementation}
All neural network architectures were implemented in TFLearn \cite{tflearn2016} with Tensorflow \cite{Tensorflow} backend.
The models were trained and tested on Ubuntu 16.04.2 LTS system with Intel Core i7-5820K CPU and Nvidia GeForce GTX 1080 8GB GPU and took between 1.5-6 hours to train depending on architecture configuration.

\section{Results}
The SPARE dataset used for this purpose has {259,200} examples in total, of which we use {153,600} for training, {28,800} for validation, and {76,800} for testing.
Our first evaluation is performed on the link counting architectures from Section~\ref{sec:linkCount} with corresponding classification accuracies provided in Table~\ref{tab:SPARERes}.
Next, link length measuring architectures from Section~\ref{sec:linkLength} are evaluated using the following error for each instance:
\begin{align}
E_{L} = \sum_{i=0}^{n} | L_i - \hat{L}_i|^2,
\label{eq:lengthEval}
\end{align}
where 
$L_i$ and $\hat{L}_i$ are the respective ground truth and predicted link length in meters.
Note that \eqref{eq:lengthEval} can be normalize by the first link length in the case of scale ambiguous training data (e.g., greyscale images from a single viewpoint).
When evaluating the end-to-end systems from Section~\ref{sec:end_to_end}, errors can arise from link counting and length estimation.
Thus, we modify error calculation \eqref{eq:lengthEval} using $L_i$ from \eqref{eq:end-end} and similarly calculate $\hat{L}_i$  as:
\begin{align}
\hat{L}_i =  \begin{cases}
\hat{l}_i, ~\text{if}~ i<=\hat{n}\\
0, ~\text{otherwise}
\end{cases}, ~i \in \{0,1,..,\hat{n}\},
\label{eq:end-endEst}
\end{align}  
where $\hat{l}_i$ is the estimated length of link $i$ and $\hat{n}$ is the estimated number of moving links.

\begin{figure}[t]
	\centering
	\includegraphics[width = 0.6\textwidth]{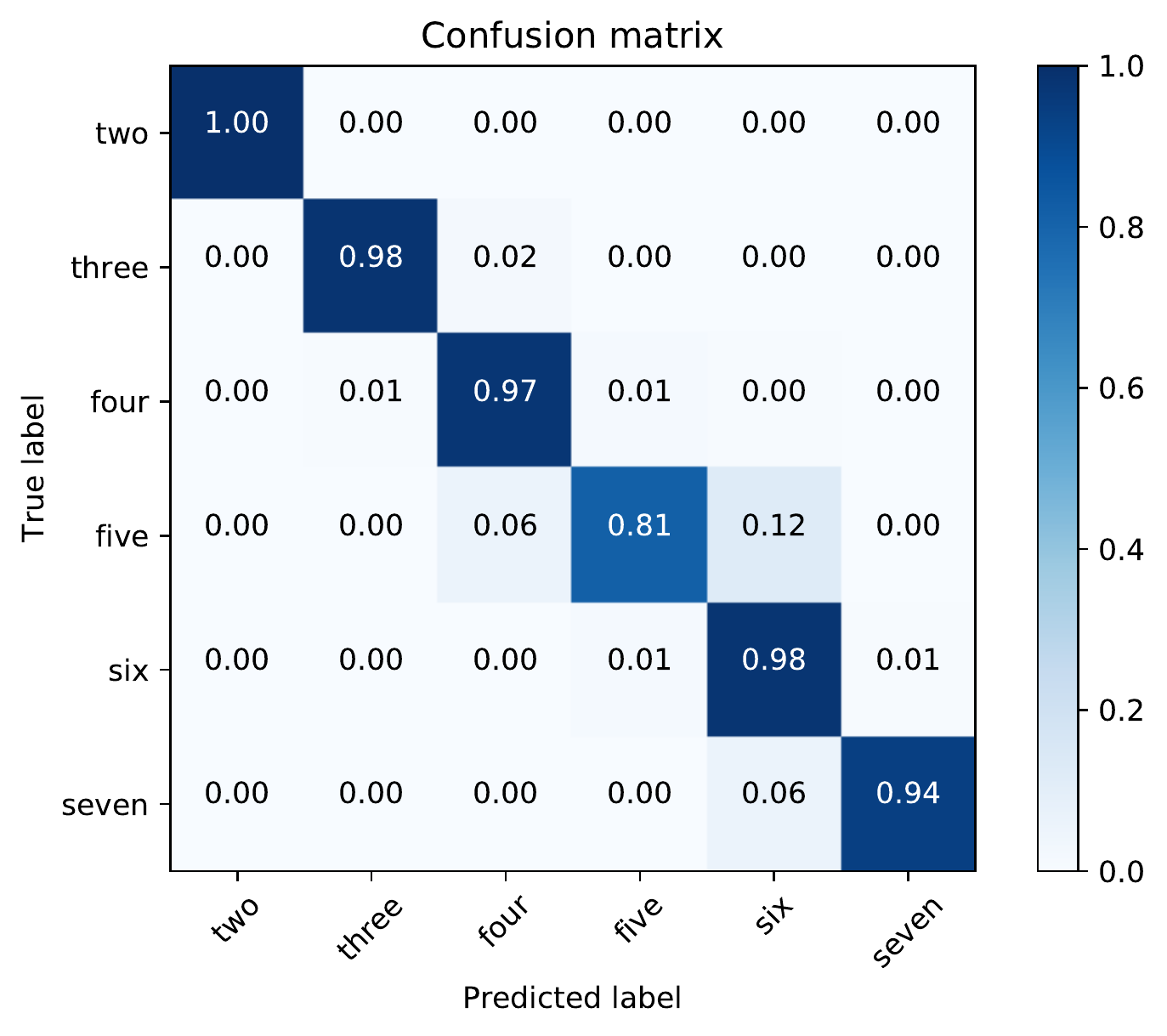}
	\caption{Confusion matrix for the link counting test evaluation of the CONV3D-Depth-MV architecture.}
	\label{fig:confConv}
\end{figure}

A summary of the SPARE test evaluation and benchmark results are provided in Table~\ref{tab:SPARERes} with further discussion in Section~\ref{sec:discuss}.
A confusion matrix in Fig.~\ref{fig:confConv} depicts the link-specific accuracy for the CONV3D-Depth-MV architecture.

\section{Discussion}
\label{sec:discuss}
The SPARE dataset is an effective tool for training multiple neural networks to generate various kinematic descriptions for articulated objects from RGBD image frames. 
Using SPARE, we generated large amount annotated data by varying the link length, color, joint angle position, and views. 
While the amount of data is sufficient to train the baseline neural networks we proposed, the SPARE framework can be used to generate additional data, in case a deeper network is required to be trained. 
For evaluation, the best overall results are for the CONV3D-Depth-MV and LSTM-Depth-MV architectures (see Table~\ref{tab:SPARERes}), which are both able to count the number of links in an articulated object with approximately 95\%. 
In addition to these results, the efficacy of multiple views over temporal information is again confirmed when the CONV3D-Depth-MV and the CONV3D-Depth-TMP input architectures are both trained and evaluated for the following two cases: 1) given that there are seven links, estimate each link lengths (Section~\ref{sec:linkLength}) and 2) count the number of links and estimate the link lengths (Section~\ref{sec:end_to_end}).
In both cases, the CONV3D-Depth-MV architecture had an order of magnitude lower average seven-link length error compared with its temporal counterpart.
Interestingly, the CONV3D-Depth-MV architecture exhibits better performance for the arguably harder problem of estimating link length and counting links (0.415~m versus 0.543~m), 
which substantiates that training end-to-end system performs better for generating kinematic description.

\section{Conclusion and Future Work}



In this work we introduced SPARE, which is the first dataset of its kind, encompassing simulated and physical articulated objects (3D-printable), and enabling limitless extensions of datasets to enable sufficient training examples for learning applications. 
Apart from training networks that learn kinematics, we also see other potential uses of this dataset, attributing to its vast number of labeled instances. 
In the field of deep learning, it is common to train networks on an extensive dataset outside of their intended application area, and then fine-tune on a more related dataset. 
Thus, given its size, we propose that pre-training RGBD-based networks on SPARE would be useful in cases where application-specific datasets have relatively few instances of training data.

In future work, we plan to extend the dataset to include set examples for tree structures and articulated objects with variable degrees of freedom for a given number links. 
Currently, in our baseline kinematic learning neural networks we use raw RGBD images; however, pre-processing the depth image and transforming input to point clouds  \cite{pointnet_Qi} or voxels  \cite{ShapeNet} could improve accuracy. 
%
Additionally, we anticipate other researchers utilizing SPARE for learning kinematic descriptions of unknown objects and for transfer learning, and we welcome feedback from the research community in regards to providing future iterations of this dataset. 

In future algorithmic work, we anticipate the use of new deep learning architectures and methods for addressing the counting of links and measuring link lengths.
Additional pre-processing of the input image, such as, converting the depth image to point cloud or adding image coordinates could also improve the prediction accuracy. 
We plan to extend our algorithm to measure the relative angle of links to generate complete Denavit-Hartenberg parameter tables for a given kinematic chain.  Finally, we will extend the complexity of the type of articulated joints and structures in SPARE.

\section*{NOTE TO REVIEWERS}
We will release the source code for SPARE and our benchmark algorithms along with the paper upon publication.

\bibliographystyle{splncs}
\bibliography{spare}
\end{document}